\pdfoutput=1
\documentclass[letterpaper]{article} 
\usepackage{aaai24}  
\usepackage{times}  
\usepackage{helvet}  
\usepackage{courier}  
\usepackage[hyphens]{url}  
\usepackage{graphicx} 
\usepackage{natbib}  
\usepackage{caption} 
\usepackage{algorithm}
\usepackage{algorithmic}
\usepackage{subfig}
\usepackage{amsmath}
\usepackage{booktabs}
\usepackage{tabularx}
\usepackage{pdfpages}

\usepackage{newfloat}
\usepackage{listings}
\DeclareCaptionStyle{ruled}{labelfont=normalfont,labelsep=colon,strut=off} 
\floatstyle{ruled}
\newfloat{listing}{tb}{lst}{}
\floatname{listing}{Listing}
\title{Spiking NeRF: Representing the Real-World Geometry by\\ a Discontinuous Representation}
\author{
    Zhanfeng Liao,
    Qian Zheng\equalcontrib,
    Yan Liu,
    Gang Pan\equalcontrib
}
\affiliations{
    \textsuperscript{\rm}Zhejiang University\\

    \{zhanfengliao, qianzheng, rliuyan2, gpan\}@zju.edu.cn
}

\usepackage{bibentry}

\begin{document}

\maketitle

\begin{abstract}
A crucial reason for the success of existing NeRF-based methods is to build a neural density field for the geometry representation via multiple perceptron layers (MLPs).
MLPs are continuous functions, however, real geometry or density field is frequently discontinuous at the interface between the air and the surface.
Such a contrary brings the problem of unfaithful geometry representation.
To this end, this paper proposes spiking NeRF, which leverages spiking neurons and a hybrid Artificial Neural Network (ANN)-Spiking Neural Network (SNN) framework to build a discontinuous density field for faithful geometry representation. Specifically, we first demonstrate the reason why continuous density fields will bring inaccuracy.
Then, we propose to use the spiking neurons to build a discontinuous density field.
We conduct a comprehensive analysis for the problem of existing spiking neuron models and then provide the numerical relationship between the parameter of the spiking neuron and the theoretical accuracy of geometry.
Based on this, we propose a bounded spiking neuron to build the discontinuous density field.
Our method achieves SOTA performance. 
The source code and the supplementary material are available at https://github.com/liaozhanfeng/Spiking-NeRF.
\end{abstract}

\section{Introduction}
3D reconstruction from RGB images is a challenging and complex task in computer vision \cite{1nerf,7neus}. 
Neural Radiance Fields (NeRF) \cite{1nerf}, a recently promising solution for novel view synthesis in an implicit manner, has also achieved very competitive results in 3D reconstruction \cite{7neus,wang2022hf,wang2023pet}.

\begin{figure*}[!t]
\centering
\includegraphics[width=1.0\textwidth]{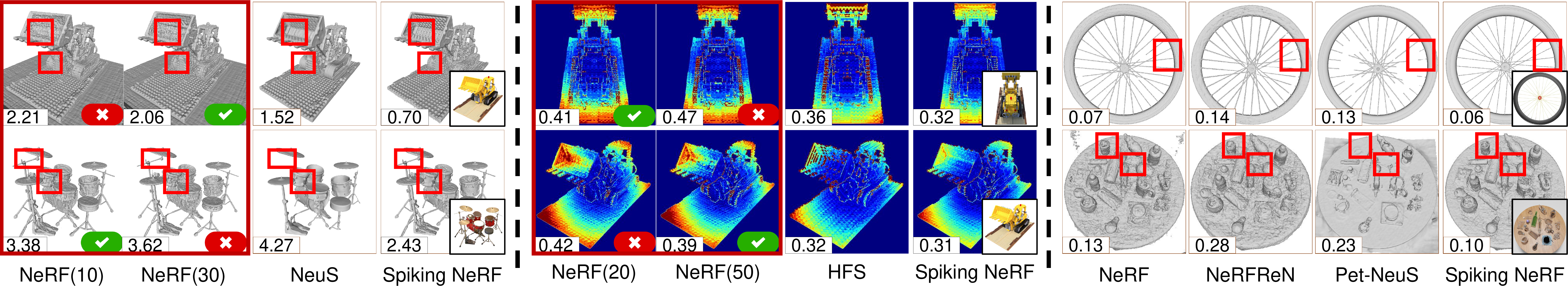} 
\caption{
Left: The extracted surfaces from NeRF. 
Each row in the first big red box represents a surface extracted by a trained NeRF using different thresholds, indicating that the optimal thresholds corresponding to different scenarios are different. 
The tick represents that the threshold is optimal. The cross represents that the threshold is not optimal.
Middle: The error maps from different views in the same scene.
Each row in the second big red box represents the depth error map of a trained NeRF's surface extracted with different thresholds from different views. It can be seen that the optimal thresholds corresponding to different views are different.
Right: The extracted surfaces from NeRF.
These figures show that the inconsistency can result in even greater errors in light density scenarios.
The image in the bottom right corner of each part represents the original image from the corresponding view. The number displayed in the bottom left corner of each image represents either the Chamfer distance (left and right) or the depth error (middle).
}
\label{intro}
\end{figure*}

One of the reasons for the success of NeRF is its ability to implicitly represent geometric information using a neural network based on conventional MLPs, which are continuous functions.
However, geometric information in the real world is discontinuous at the interface between the air and the surface, which is inconsistent with the computational representation in conventional MLPs.
This inconsistency poses three problems.
\textbf{Post-processing}. 
Existing methods require a post-processing approach (i.e., filtering with a threshold \cite{30peng2021neuralbody,31dnerf,32boss2021nerd}) to extract discontinuous geometric representations from learned continuous fields.
However, determining the optimal threshold requires empirical and tedious tuning, and many methods rely on manual selection of the threshold for different scenarios \cite{1nerf,31dnerf,32boss2021nerd} (see the left of Fig. \ref{intro}).
\textbf{Optimal threshold perturbation}. 
NeRF-based methods optimize the network view by view, resulting in the optimal threshold of the surface to perturbate (see supplementary material for more details).
However, existing methods use the same threshold to filter output values \cite{1nerf,30peng2021neuralbody,31dnerf,32boss2021nerd}, which cannot obtain accurate geometric information because the optimal filtering thresholds under different views vary (see the middle of Fig. \ref{intro}).
\textbf{\textit{Light density scenarios}}\footnote{These scenarios contain several rays, along which the density with non-zero values distribute within a very narrow range (e.g., thin objects) or with small values (e.g., semi-transparent objects). In real life, semi-transparent objects (e.g., windows, glasses) and thin objects (e.g., branches, axles, nets) can be seen everywhere.}.
The inconsistency can result in even greater errors in light density scenarios (see the right of Fig. \ref{intro}).
Because it is difficult for continuous functions like conventional MLPs to generate quite different densities (i.e., significantly different outputs) for points close to each other (i.e., similar inputs), resulting in the disappearance of the object.

Although there are some attempts to address the aforementioned problems, these methods are still based on continuous representation and do not provide a fundamental solution to those problems.
For example, some works replace the density field with other fields to avoid threshold selection (e.g., \cite{5unisurf,7neus,yariv2021volume}).
However, these methods cannot use the filter value to get an accurate geometric representation (see the left of Fig. \ref{intro}). 
Moreover, they fail on light density scenarios (see the right of Fig. \ref{intro}).
Some works modify the input to improve its frequency or discretization, thereby mitigating the inconsistency between continuous and discontinuous representations.
However, these methods still fail on light density scenarios \cite{wang2022hf,wang2023pet} (see the right of Fig. \ref{intro}).
Some works are designed for specific scenarios (e.g., \cite{29guo2022nerfren,levy2023seathru}).
However, these methods have poorer performance in general scenarios.
Moreover, they mainly focus on novel view synthesis, and the quality of the 3D reconstruction is low (see the right of Fig. \ref{intro}).

Compared to the ANN transmitting continuous values, the SNN transmits discontinuous spikes.
The discontinuity of the SNN is expected to represent the discontinuous geometric information in the real world and fundamentally resolve the inconsistency between real-world representation and computational representation in conventional MLPs.
Moreover, the threshold of spiking neurons in SNNs can serve as a filtering threshold through a parameter learning scheme (e.g., \cite{21li2022brain,22SpikingTransformers}), which avoids the issue of manually specifying the threshold. 
Different from traditional cognitive applications \cite{21li2022brain,22SpikingTransformers,han2023symmetric,zou2022towards}, using the spiking representation to model 3D geometry from a numerical perspective is a non-cognitive application \cite{aimone2022review,ren2023spiking}.

In this paper, we address the problems of continuous geometric representation in the conventional NeRF by proposing spiking NeRF, which is based on a hybrid ANN-SNN framework to model the real-world 3D geometry in a discontinuous representation.
First, we build the relationship of the spiking threshold, maximum activation, and depth error for our discontinuous computational representation.
Second, based on this relationship, we observe that when the spiking threshold is sufficiently large, the depth error is sufficiently small.
However, the spiking threshold cannot be set sufficiently high in semi-transparent scenarios and cannot be set to infinity in practical implementation.
We further discover a way to maintain a small error under a finite spiking threshold, which is to control the maximum activation.
Last, based on our analysis, we propose the Bounded Full-precision Integrate and Fire (B-FIF) spiking neuron to build a hybrid ANN-SNN framework.
Meanwhile, we design a corresponding training pipeline for the hybrid ANN-SNN framework and verify the effectiveness on mainstream datasets and light density scenarios.
Our contributions can be summarized as follows:

\begin{itemize}
     \item 
     We build the relationship of the spiking threshold, maximum activation, and depth error. 
     Moreover, we constrain the bound of depth error by the spiking threshold and maximum activation. 
     This bound can be leveraged to facilitate the real-world application for geometric estimation without knowing the ground truth.
     \item 
     We propose the B-FIF spiking neuron based on the aforementioned analysis and build a hybrid ANN-SNN framework. 
     Meanwhile, we design the corresponding training pipeline and training strategy for the hybrid ANN-SNN framework. 
     We verify the effectiveness on mainstream datasets and light density scenarios.
\end{itemize}
\section{Related Work}

\subsection{Neural Implicit Representations}
Recently, neural implicit functions have emerged as an effective representation of 3D geometry \cite{58atzmon2019controlling,65peng2020convolutional,66saito2019pifu,67xu2019disn} and appearance \cite{68liu2020neural,69liu2020dist,75schwarz2020graf,76sitzmann2019scene} as they represent 3D content continuously and without discretization while simultaneously having a small memory footprint.
Most of these methods require 3D supervision. 
However, several recent works \cite{71niemeyer2020differentiable,79yariv2020multiview} demonstrated differentiable rendering for training directly from images.
Some works use estimated depth information for surface rendering without pixel-accurate object masks \cite{11refnerf}. 
Some works enhance the accuracy of geometric information by incorporating point cloud information and warp operations \cite{fu2022geo,darmon2022improving}.
Some works do not model the density field and use other fields to avoid threshold selection \cite{5unisurf,7neus,fu2022geo,darmon2022improving,yariv2021volume}.
However, these works are still based on a continuous representation, the same threshold cannot accurately divide the surface.
Different from previous methods, we model the 3D geometric information in a discontinuous representation by proposing a hybrid ANN-SNN framework.

\subsection{Spiking Neural Networks in Computer Vision}
Over the past few years, brain-inspired \cite{18li2021fast} deep SNNs using spike trains \cite{17isspikingsafe} have gradually replaced ANNs in various tasks \cite{zhang2023defects,zhang2023predicting,ororbia2023spiking}, and it is extensively utilized to develop energy-efficient solutions for various tasks. 
Some works use SNNs to process video or event streams \cite{23zhu2022event,gehrig2020event}.
Some works exploit the discontinuity of the network to enhance its robustness \cite{25adversarial,27inherentadversarial}. 
Some works leverage the neural parameter of SNNs to further improve the utilization of available information \cite{21li2022brain,22SpikingTransformers}. 
Although previous works have demonstrated SNN applications on a wide range of tasks, they are still limited in their performance \cite{li2021free,fang2021deep}. 
Meanwhile, there has been a growing interest in exploring the potential benefits of combining ANNs and SNNs \cite{kugele2021hybrid,liu2022enhancing}. 
By combining ANNs and SNNs, better performance can be achieved while reducing the time step of SNNs.
Different combination strategies have been explored for a variety of tasks.
A group of work employs the strategy of processing the accumulated spike train of SNNs with ANNs \cite{kugele2021hybrid,lee2020spike}.
In these works, the SNN is used as an efficient encoder of spatio-temporal data. The output of the SNN is accumulated to summarize the temporal dimension before the ANN processes the accumulated features \cite{kugele2021hybrid,lee2020spike}. 
A second line of work uses a strategy where the output of the independently operating SNN and ANN is fused \cite{lele2022bio,zhao2022framework}.
In these works, the outputs of both networks are fused based on heuristics \cite{lele2022bio} or accumulation based on the output of the ANN \cite{zhao2022framework}. 
Different from the previous methods, we apply the hybrid ANN-SNN framework to a non-cognitive application.

\section{Preliminary}
\subsection{Neural Radiance Fields}
NeRF \cite{1nerf} represents a scene as a continuous volumetric field, where the density $\sigma \in R$ and radiance $c\in R^3$ at any 3D position $\textbf{x}\in R^3$ under viewing direction $\textbf{d}\in R^2$ are modeled by a MLP $f_\theta: (\textbf{x}, \textbf{d})\to (c, \sigma)$, with $\theta$ as learnable parameters. To render a pixel, the MLP first evaluates points sampled from the camera ray $r = \textbf{o} + t\textbf{d}$ to get their densities and radiance. The light starts at 0 and stops at a hypothetical large density $T$.
Define $C_\text{p}$ as the estimated pixel color and $\hat{C_\text{p}}$ as the ground truth. NeRF is optimized by minimizing the photometric loss:
\begin{equation}
    \label{eq:2}
    L_\text{rgb}=\sum_\text{p}\left|\left|C_\text{p} - \hat{C_\text{p}}\right|\right|.
\end{equation}
Here, $\text{p}$ refers to each pixel.
There are currently two main methods for depth estimation: one based on the integration \cite{12nerfactor,14Depth-supervisedNeRF} and the other based on the threshold to get global shape \cite{7neus, 20dexnerf}. 
Since we focus on a global geometric shape extracted from the network by threshold rather than a depth map dependent on perspective, this paper mainly follows the second method to define the depth.

\subsection{Spiking Neuron}
\paragraph{Integrate-and-fire model.}
We introduce the well-known Integrate-and-Fire (IF) model \cite{dayan2005theoretical}.
Given a membrane potential $u_{t}$ at time step $t$, the membrane potential $u^\text{pre}_{t+1}$ before firing at time step ${t+1}$ is updated as:
\begin{equation}
    \label{eq:4}
    u^\text{pre}_{t+1} = u_{t} + Wx_{t+1}.
\end{equation}
Here, $W$ and $x_{t+1}$ represent respectively the weight and the output from the previous layer at time step ${t+1}$.
${V_\text{th}}$ is the firing threshold. 
The spiking neuron will fire a spike when $u^\text{pre}_{t+1}$ exceeds ${V_\text{th}}$ (see Fig. \ref{bound}), and then $u_{t+1}$ resets to $0$. 
The spike output is given by:
\begin{equation}
    \label{eq:5}
    o_{t+1} = 
    \begin{cases}
        0&u^\text{pre}_{t+1}<V_\text{th} \\
        1&\text{otherwise},
    \end{cases}
\end{equation}
\begin{equation}
    \label{eq:6}
    y_{t+1} = o_{t+1} \cdot V_\text{th}.
\end{equation}
After firing, the spike output ${y_{t+1}}$ will propagate to the next layer and become the input ${x_{t+1}}$ of the next layer. Note that we omit the layer index for simplicity.
\paragraph{Full-precision integrate-and-fire model.}
However, the spikes convey less information than the floating-point number, and networks using IF do not perform well in regression tasks \cite{ranccon2022stereospike,lu2021learning,shrestha2018slayer}.
It’s difficult to ensure accuracy by directly using IF for modeling. 
To obtain full-precision information, \cite{21li2022brain} changes the Eqn. \ref{eq:6} to:
\begin{equation}
    \label{eq:7}
    y_{t+1} = o_{t+1} \cdot u^\text{pre}_{t+1}.
\end{equation}
By replacing the output $V_\text{th}$ with $u^\text{pre}_{t+1}$, the full-precision information is reserved. Due to the full-precision information meeting our requirements, we will only consider one time step and omit $t$ in the following formulations.
\section{Spiking NeRF}

To make the density field discontinuous, a trivial way is to introduce spiking neurons in NeRF, which replaces the last activation layer of the density network with spiking neurons (e.g., IF, FIF). 
However, on the one hand, as previously mentioned, using IF is difficult to ensure accuracy, and on the other hand, directly replacing the activation layer with FIF will result in a significant depth error when the spiking threshold is small (see ablation study for more details).

In this section, we first analyze what kind of spiking neurons can solve problems of post-processing and optimal threshold perturbation, and alleviate the problem of light density scenarios by building the relationship of the spiking threshold, maximum activation, and depth error. 
Then, based on this relationship, we further propose our method.
\subsection{Relationship between Parameters of Spiking Neuron and Depth Error}
We define that $d_\text{v}$ is the estimated depth and $d$ is the accurate depth. 
$T$ refers to the length of the sampling range and $\Delta t$ refers to the sampling interval. 
$V_\text{th}$ is the spiking threshold and $V_\text{max}$ is the maximum density value. 
For a well-trained NeRF, the $V_\text{th}$ is also the non-zero minimum value of its density field, and the density value of the first point that a ray encounters with a non-zero density is $V_\text{th}$.
We build the relationship based on the following Proposition 1 (see supplementary material for proof).
\paragraph{Proposition 1.} 
For a well-trained NeRF, we have:
\begin{equation}
    \resizebox{.89\hsize}{!}{$\left |d - d_\text{v}\right |<\max((\Delta t - Te^{-V_\text{max}\Delta t}) e^{-V_\text{th}\Delta t}, T(1-e^{-V_\text{max}T})e^{-V_\text{th}\Delta t})$}.
\end{equation}

%
Based on the relationship, when $V_\text{th}$ is sufficiently large, $e^{-V_\text{th}\Delta t}$ is sufficiently small, resulting in a sufficiently small error.
Meanwhile, for a fixed $V_\text{th}$, a small $V_\text{max}$ can decrease the error (see supplementary material for more details).

\begin{figure}[!t]
\centering
\includegraphics[width=1\columnwidth]{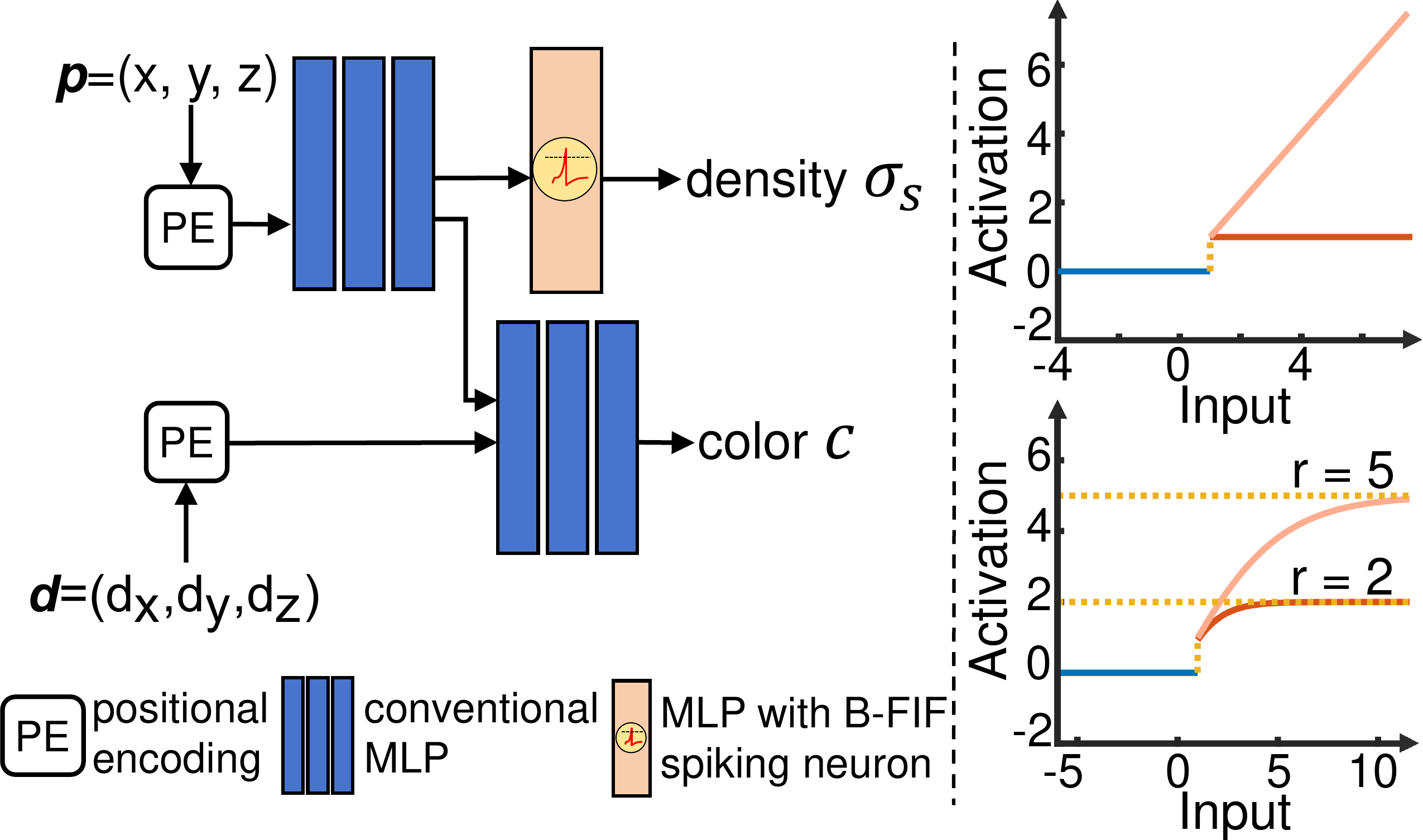}
\caption{Framework overview of spiking NeRF and an illustration of different existing spiking neuron models and the proposed one. Left: The network structure of our approach. We use a NeRF model following \cite{1nerf} but excluding the last activation layer of the density network. Instead of using ReLU, we use B-FIF spiking neurons to make the density field discontinuous. Right top: the IF and FIF. Right bottom: B-FIF with different $r$ ($r=2$ and $5$). These curves show that B-FIF becomes more similar to FIF as the parameter $r$ increases. And when the $r$ is sufficiently large, B-FIF degenerates to the FIF.}
\label{bound}
\end{figure}

However, $V_\text{th}$ cannot be set sufficiently high in semi-transparent scenarios and cannot be set to infinity in practical implementation.
Meanwhile, FIF does not have $V_\text{max}$ constraint (see the right of Fig. \ref{bound}), resulting in a large positive $(\Delta t - Te^{-V_\text{max}\Delta t})$. 
In this case, significant errors may occur in semi-transparent scenarios and practical implementation. 
Direct use of FIF cannot completely solve the above-mentioned problems.
Therefore, $V_\text{max}$ should have a constraint to decrease the error and we need a spiking neuron with a relatively small maximum activation.
\paragraph{B-FIF: Bounded full-precision integrate-and-fire spiking neuron.}
Based on previous analysis, the maximum activation close to the spiking threshold can ensure lower error. 
Therefore, we need to constrain the maximum activation of spiking neurons to decrease the error.
In this paper, we use $\tanh()$ to constrain the maximum activation (see supplementary material for other bounded functions).
The $u^\text{pre}$ in Eq. \ref{eq:7} is reformulated as:
\begin{equation}
    \label{eq:9}
    u^\text{pre} = k\cdot\tanh({u+Wx}),
\end{equation}
\begin{equation}
    \label{eq:10}
    \sigma = y = o \cdot u^\text{pre}.
\end{equation}

Here, $k$ is a learnable parameter.
By increasing $k$, it is ensured that the spiking neuron can have a larger spiking threshold.
The proposed neuron constrains the maximum output value while ensuring accuracy as it is also a full-precision neuron. 
So the spiking neuron can ensure that the network is trained accurately and also decrease errors when the spiking threshold is relatively small as its maximum activation approaches its spiking threshold.


\subsection{Hybrid ANN-SNN Framework}
\paragraph{B-FIF implementation.}
Directly using the previously mentioned spiking neuron can lead to slow training. It is potentially caused by the derivative of the initial network output approaching $0$ (see Fig. \ref{bound}), leading to a smaller learning step size. Therefore, to better train the network, we introduce a learnable parameter $r$ for proposed neurons to increase the derivative of the initial network output (see Fig. \ref{bound}), which can improve training.
The $u^\text{pre}$ in Eq. \ref{eq:9} is reformulated as:
\begin{equation}
    \label{eq:20}
    u^\text{pre} = k\cdot r\cdot\tanh({u+Wx\over r}).
\end{equation}
We set the initial value of $r$ and $k$ to $100$ and $1$ to maintain similarity to the original NeRF output (i.e., ReLU \cite{1nerf,6RGB-Dsurfacereconstruction,7neus}) near point $0$ (see Fig. \ref{bound}). 
This setting ensures that the network maintains performance similar to NeRF during the initial training stage, which can ensure that the range of the derivative approaching $1$ is expanded. 
In addition to the above advantage, this initial strategy and the proposed neuron in Eq. \ref{eq:20} have another advantage the network is trained with an increased penalty for small density regions that should not appear, and it can limit the density value that should not appear above the spiking threshold more easily during training. 

When $r$ is sufficiently large, the B-FIF degenerates into the FIF, and when the maximum activation approaches the threshold, the B-FIF degenerates into the IF (please find Fig. \ref{bound} for the comparison of different spiking neuron models).
\paragraph{Loss function.}
Based on Proposition 1, it is necessary to have a large spiking threshold to obtain accurate geometric information. The network can learn the spiking threshold during the training process. However, due to the difficulty of the network spontaneously pushing the spiking threshold to a larger value, the spiking threshold at the end of the training process may not be relatively large. Therefore, we propose a regularization term $L_\text{v}$ to increase the spiking threshold and ensure that the spiking threshold does not remain a small value after the training is completed. The $L_\text{v}$ is as follows:
\begin{equation}
    L_\text{v}={1\over V_\text{th}}.
\end{equation}


Following \cite{11refnerf}, we use a regularization term $L_\text{g}$ to maintain the smoothness of the geometric representation in the initial stage, improving the network convergence. 
 The $L_\text{g}$ is as follows:
\begin{equation}
L_\text{g} = \sum_\text{p} \sum_{i} w_{i} \max(-\textbf{d}_\text{p}\nabla\sigma_{i}, 0).
\end{equation}
Here, $\text{p}$ refers to each pixel as defined in \cite{29guo2022nerfren}.
$i$ refers to each sampling point and $w_{i}$ refers to the weight at each sampling point as defined in \cite{11refnerf}.
However, in \cite{11refnerf}, they specifically designed a MLP to predict the normal vector, which generated more parameters that needed to be learned. 
We directly use the normal vector calculated by the gradient of density on the input coordinates to calculate $L_\text{g}$, reducing the number of parameters while reducing network complexity.

Finally, We optimize the following loss function:
\begin{equation}
    \label{eq:11}
    L=L_\text{rgb}+\lambda_1 L_\text{v}+\lambda_2 L_\text{g}.
\end{equation}

\paragraph{Surrogate gradient.}
The non-differentiability of the firing function remains one of the most significant challenges in training SNNs \cite{neftci2019surrogate}.
Direct training requires the use of the surrogate gradient \cite{16Deeplearningwithspikingneurons,21li2022brain,28li2021differentiable,25adversarial}.
%
There are 2 kinds of surrogate gradient, the surrogate gradient for conventional spiking neurons and the surrogate gradient for full-precision spiking neurons.
The proposed neuron is based on the full-precision spiking neuron, so we use the surrogate gradient for full-precision spiking neurons.
However, there are multiple time steps in \cite{21li2022brain}, and the surrogate gradient of the spiking threshold is based on the number of spikes, while our method only uses one time step. 
Directly using the surrogate gradient in \cite{21li2022brain} can lead to unstable training (see supplement material for more details).
In this paper, we use the piecewise linear function similar to \cite{28li2021differentiable} as the surrogate gradient for $V_\text{th}$ to avoid the gradient that is $0$ almost everywhere. The surrogate gradient of B-FIF is as follows:
\begin{equation}
    \label{eq:15}
    {\partial y\over\partial u^\text{pre}} = o,
\end{equation}
\begin{equation}
    \label{eq:16}
    {{\partial y}\over{\partial V_\text{th}}} = \lambda \max(0,{k-\left|u^\text{pre} - V_\text{th}\right|\over k^2})u^\text{pre}.
\end{equation}
Meanwhile, due to the use of full-precision spiking neurons, we only need one time step to meet requirements. 
So theoretically, our training complexity is similar to that of ANN. 

\paragraph{Training strategy.}
During the initial stage of network training, the density field exhibits significant deviations.
A larger spiking threshold will render the network untrainable because most of the membrane potential cannot reach the spiking threshold and is not in the range of surrogate gradient non-zero. 
So we initially set the spiking threshold to $0$. Furthermore, we set $\lambda_1$ to $0.15$ at first and continue to increase $\lambda_1$ as the training process progresses.
To preserve true high-frequency geometric information, we initially set the $\lambda_2$ to $0.0001$ and gradually decrease the proportion of $L_\text{g}$ in $L$ as the training progresses by decreasing $\lambda_2$.
\begin{figure*}[htb]
\centering
\includegraphics[width=1\textwidth]{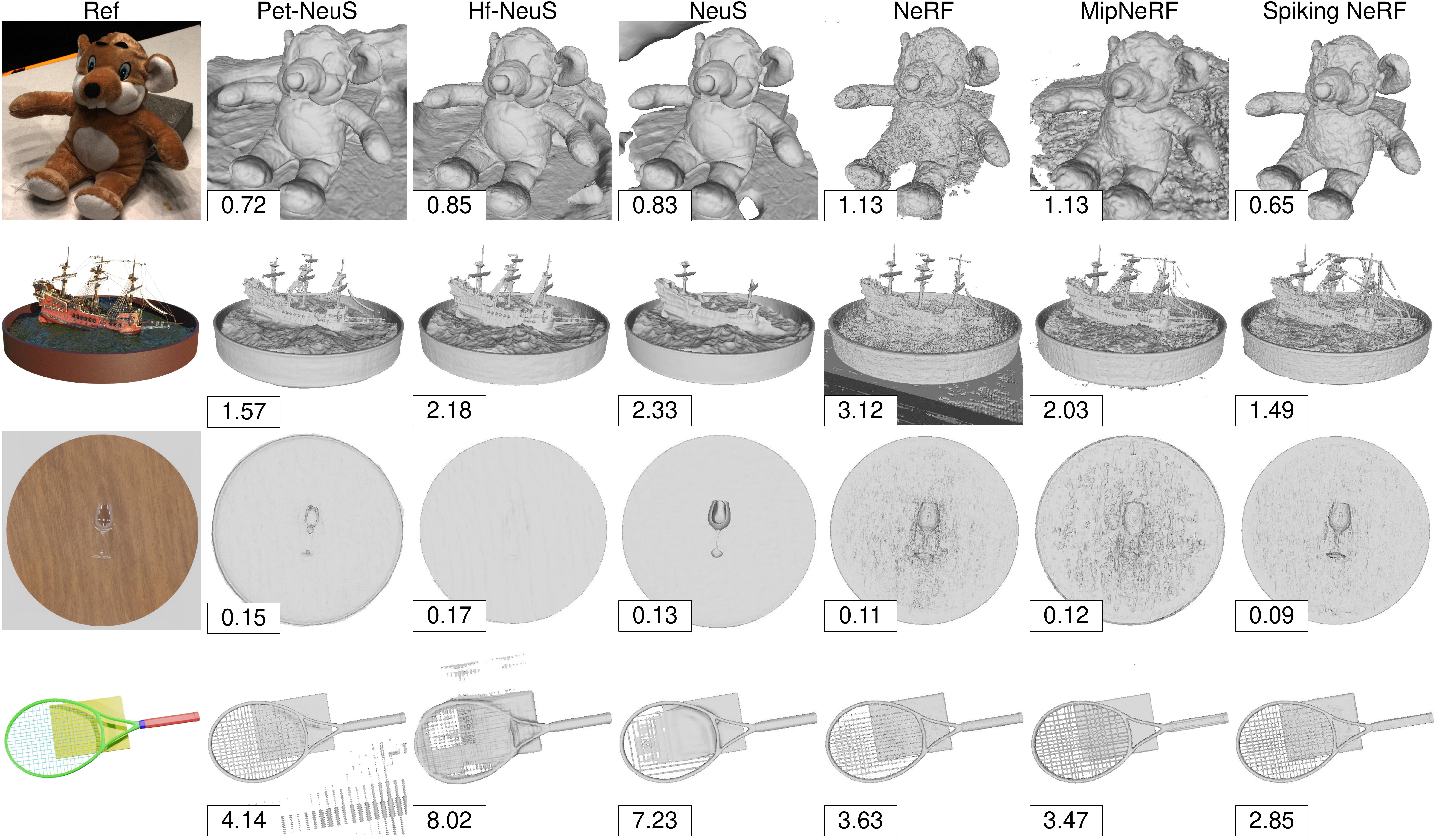}
\caption{Visual Quality Comparisons on surface reconstruction on Blender dataset \cite{1nerf}, DTU dataset \cite{80jensen2014large}, semi-transparent dataset \cite{20dexnerf}, and thin object dataset.
We show the Chamfer distance in the bottom left corner of the image. The results of the 2nd and 4th rows are multiplied by $10^2$.
}
\label{Blender1}
\end{figure*}
\section{Experiments}
\subsection{Experimental Settings}
\paragraph{Datasets.}
We evaluate our methods in 8 scenes from the Blender dataset \cite{1nerf} and 6 scenes from Dex-NeRF \cite{20dexnerf}. 
Following most 3D reconstruction methods \cite{7neus, wang2022hf, wang2023pet,5unisurf}, we also evaluate our methods in the DTU dataset \cite{80jensen2014large}. 
These datasets cover different types of scenes or objects and are benchmarks for NeRF-based methods.
Furthermore, we additionally use 2 self-created scenes to show that our method is better than previous methods in special scenarios (see supplementary material for more details).

\paragraph{Implementation details.}
We sample 1024 rays per batch and train our model for 400k iterations for 10 hours on a single NVIDIA RTX3090 GPU. 
Our network architecture and initialization scheme are similar to those of NeRF \cite{1nerf} and we model the background using NeRF++ \cite{2nerf++} with the same settings as NeuS in the real scene dataset. 
\paragraph{Metrics.}
Following most 3D reconstruction methods, we measure the reconstruction quality with the Chamfer distances \cite{7neus,wang2022hf,wang2023pet,fu2022geo,darmon2022improving}.
\subsection{Overall Performance}
\paragraph{Comparison methods.}We compared our method with 2 kinds of NeRF-based methods,
conventional NeRF-based methods and SDF-based methods.
For conventional NeRF-based methods, we compared with NeRF \cite{1nerf} and MipNeRF \cite{8mipnerf} (referred to as ``Mip" in the table).
For SDF-based methods, we compare with NeuS \cite{7neus}, HF-NeuS \cite{wang2022hf} (referred to as ``HFS" in the table), and PET-NeuS \cite{wang2023pet} (referred to as ``PET" in the table). 
PET-NeuS and HF-NeuS are follow-up works on NeuS and achieve more detailed geometry reconstruction. 
We did not include IDR \cite{79yariv2020multiview}, UNISURF \cite{5unisurf}, or VolSDF \cite{yariv2021volume} as NeuS had shown superior performance on Chamfer distances.
We report the Chamfer distances in Tab. \ref{table3}, Tab. \ref{table2}, and Tab. \ref{table1}, and conduct qualitative comparisons in Fig. \ref{Blender1}. 
\paragraph{Quantitative comparison.}
These SDF-based methods failed in semi-transparent scenes (shown in Tab. \ref{table3}), resulting in significant Chamfer distances. 
This indicates that the SDF-based method cannot reconstruct semi-transparent scenes.
Meanwhile, as shown in Tab. \ref{table2}, our average Chamfer distances are the lowest in the Blender dataset \cite{1nerf}. 
This indicates that our method can obtain more accurate geometric information. 
It is rare \cite{ren2023spiking} that the result of SNN is better than that of ANN.
As shown in Tab. \ref{table1}, we have a similar average Chamfer distance to SDF-based methods in the DTU dataset.
While our method achieves leading results, some of our results are less competitive compared to some mainstream methods (e.g. PET-NeuS, HF-NeuS). 
Because SDF-based methods adopt lower frequency position encoding and constrain the variation of output with input, they have good performance for low-frequency and smooth scenes. 
However, when these premises are not met, their performance will decrease.

\begin{table}[t]
\Huge
    \centering
    \resizebox{1.03\hsize}{!}{
    \begingroup
    \fontsize{45}{50}\selectfont
    \begin{tabular}{l|c|c|c|c|c|c|c|c|c}
    \bottomrule
          & wheel &tennis  &mount  &glass  &turbine  &clutter  &pawn    &pipe    & Avg. \\
        \midrule
        PET   & 13.23  &4.14    &15.94  &15.32  & 15.34   & 23.28   &14.31   &14.23   & 14.47\\ 
        HFS   & 30.41  &8.02    &15.35  &17.24  & 11.53   & 17.23   &15.78   &12.57   & 16.01 \\ 
        NeuS  & 23.85  &7.23    &12.75  &13.14  & 11.70   & 25.32   &14.30   &11.53   & 14.97 \\ 
        NeRF  & 7.47   &3.63    &10.22  &11.53  & 10.06   & 13.15   &12.04   &10.12   & 9.77 \\ 
        Mip   & 11.31  &3.47    &10.30  &12.74  & 10.25   & 12.36   &11.67   &\textbf{9.23}    & 10.16 \\ 
        Ours  & \textbf{6.42}   &\textbf{2.85}    &\textbf{9.87}   &\textbf{9.45}   & \textbf{9.82}    & \textbf{10.44}   &\textbf{10.87}   &10.32   & \textbf{8.75} \\
        \bottomrule
    \end{tabular}
    \endgroup
    }
      \caption{Quantitative Comparison on light scenarios \cite{20dexnerf}. We show the Chamfer distance $\times 10^{-2}$ for the reconstructed geometry on 6 scenes from the Dex-NeRF dataset and 2 scenes from our Blender dataset.}
  \label{table3}
\end{table}

\begin{table}[t]

    \centering
    \resizebox{1\hsize}{!}{
    \begingroup
    \fontsize{30}{35}\selectfont
    \begin{tabular}{l|c|c|c|c|c|c|c|c|c}
    \bottomrule
      & Lego &Chair &Mic   &Ficus &Hotdog&Drums &Mats&Ship   & Avg. \\
    \midrule
    PET    &\textbf{0.58}  &\textbf{0.65}  &\textbf{0.59}  &0.71  &1.02  &2.53  & \textbf{1.05}    & 1.57  &1.09\\ 
    HFS    &0.96  &0.65  &0.72  &0.87  &1.35  &3.82  & 1.08    & 2.18  &1.45 \\ 
    NeuS   &1.52  &0.70  &0.85  &1.67  &1.40  &4.27  & 1.08    & 2.33  &1.73 \\ 
    NeRF   &2.06  &0.75  &0.95  &0.56  &1.83  &3.38  & 1.12    & 3.12  &1.72 \\ 
    Mip    &1.92  &0.90  &1.13  &0.55  &1.98  &3.34  & 1.30    & 2.03  &1.64 \\ 
    Ours   &0.70  &0.66  &0.72  &\textbf{0.54}  &\textbf{0.94}  &\textbf{2.43}  & 1.10    & \textbf{1.49}  &\textbf{1.07} \\
    \bottomrule
    \end{tabular}
    \endgroup
    }
    \caption{Quantitative Comparison on Blender. We show the Chamfer distance $\times 10^{-2}$ for the reconstructed geometry on 8 scenes from the Blender dataset. }
  \label{table2}
\end{table}

Since our method is based on the conventional NeRF and is similar to NeRF, the robustness of our method closely resembles that of NeRF. 
The actual performance can be seen in Tab. \ref{table2} and Tab. \ref{table1} which show results on the general real scene (i.e. DTU dataset \cite{80jensen2014large}) and the general synthetic scene (i.e. Blender dataset \cite{1nerf}). 
For conventional NeRF-based methods, the results obtained by manually selecting the optimal threshold are worse than our methods.
Because our method uses the hybrid ANN-SNN framework to model the 3D geometric information in a discontinuous representation, the reconstructed geometric information is more accurate.

\paragraph{Visual quality comparison.}
As shown in the 5th and 6th columns of Fig. \ref{Blender1}, with conventional NeRF-based methods, due to the lack of surface constraints, the reconstruction results have many erroneous high-frequency information.
Additionally, there is a loss of reconstruction accuracy for thin objects and semi-transparent scenes. 
As shown in the 2nd, 3rd, and 4th columns of Fig. \ref{Blender1}, with SDF-based methods, the reconstruction resulted in significant errors in high-frequency changes in the image, such as failing to reconstruct holes that should have been present and surfaces that should have been separate. 
Moreover, NeuS demonstrated excessive smoothness in the 2nd row of Fig. \ref{Blender1}. NeuS, HF-NeuS, and PET-NeuS fail in the reconstruction of semi-transparent scenes. 
Even though HF-NeuS uses some methods to reconstruct the high-frequency information, it still fails in some thin high-frequency positions. 
Because our method uses the hybrid ANN-SNN framework to model the 3D geometric information in a discontinuous representation, the reconstructed geometric information is more accurate.

\begin{figure}[t]
\centering
\includegraphics[width=1\columnwidth]{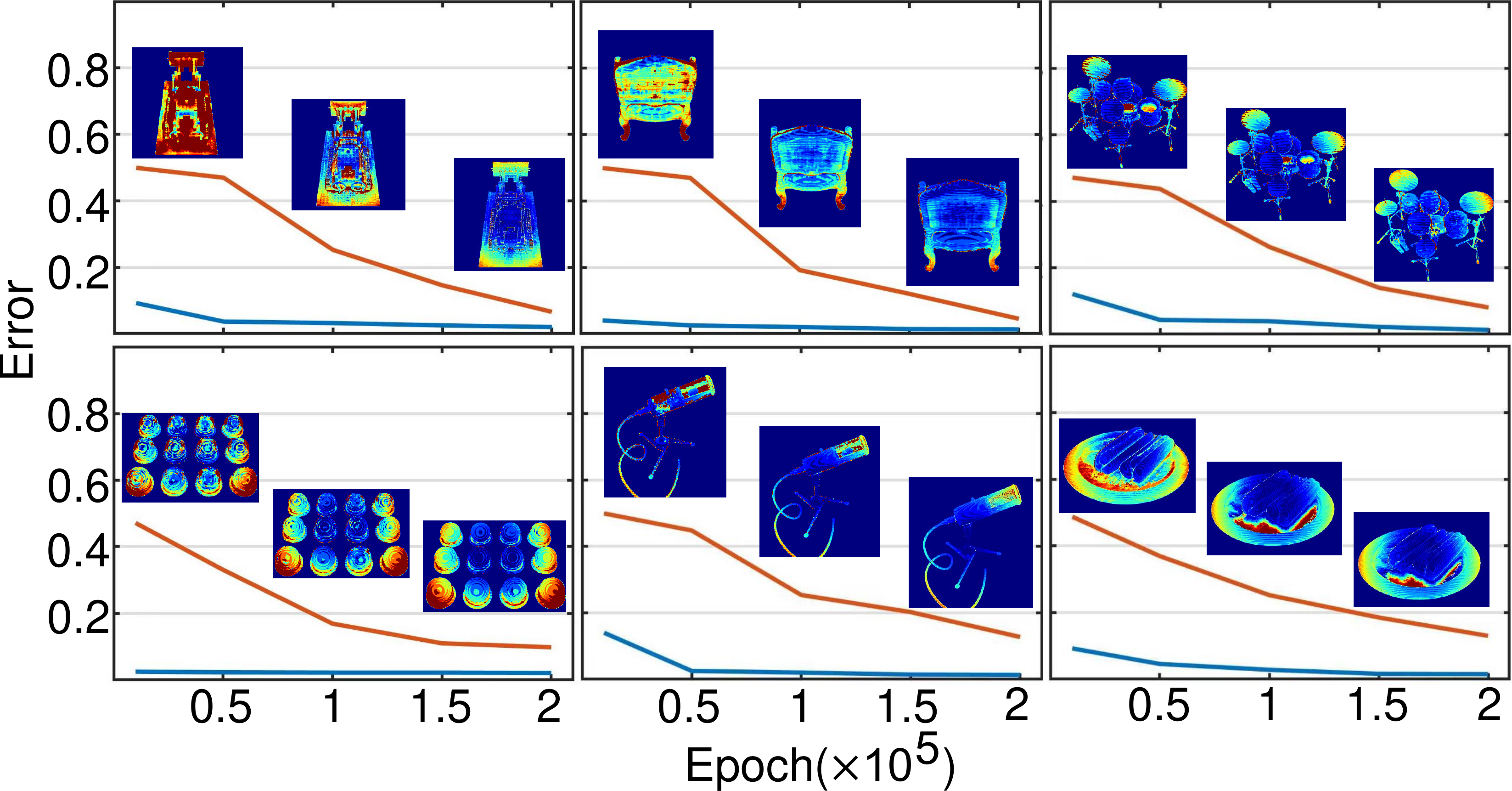}
\caption{The relationship between the upper bound and the average depth error during training. We show 6 scenes from Blender dataset \cite{1nerf}. We randomly choose a view for displaying from each scene, and compute error and upper bound in $\text{epoch}=10\text{K}$, $50\text{K}$, $100\text{K}$, $150\text{K}$ and $200\text{K}$. The red curve represents the upper bound while the blue curve represents the average depth error during training. It can be seen that the average depth error decreases with the upper bound and the average depth error keeps being less than the upper bound during training.}
\label{threshold}
\end{figure}

%
\subsection{The Practicality of Upper Bound of Proposition 1}
To explore whether the bound is practical, we verify whether there is a correlation between the bound and the average depth error and whether the bound has similar patterns across different scenarios. 
Because the Blender dataset provides depth information, we conducted our experiments on it.
As shown in Fig. \ref{threshold}, the average depth error is under the bound. 
Moreover, there is indeed a correlation between the bound and the average depth error, and the bound has similar patterns across different scenarios. 
It means that the bound can be leveraged to facilitate real-world applications for geometric estimation without knowing the ground truth. However, note that there is still a relatively large gap between the bound and the average depth error. 
While we made a new attempt in the field of NeRF and this bound can be used to a certain extent, it is still not enough to be practical.
\subsection{Ablation Study}
\begin{figure}[t]
\centering
\includegraphics[width=0.95\columnwidth]{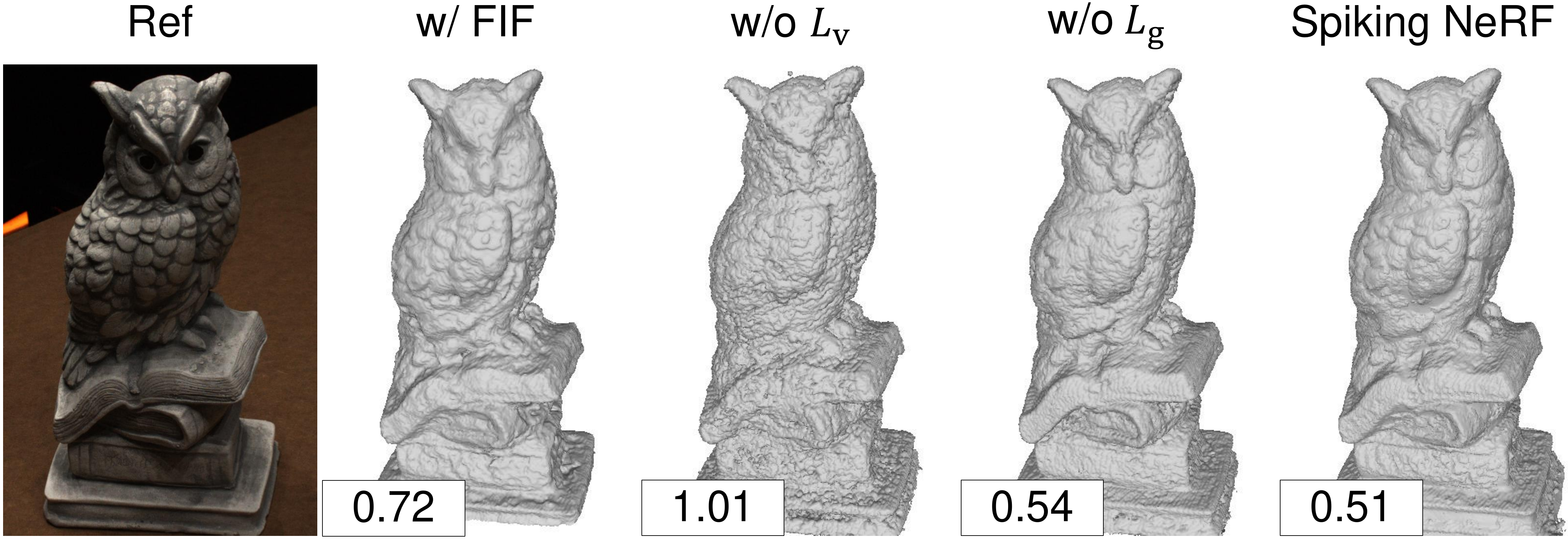}
\caption{Ablation studies. 
We show qualitative results and report the quantitative metrics in Chamfer distance.
}
\label{ablation}
\end{figure}

\paragraph{Validation for proposition 1.}
When $L_\text{v}$ is minimized, the spiking threshold will continue to increase, and when $L_\text{v}$ is removed, the spiking threshold of the final network will remain at a lower value. 
To validate proposition 1, we design an experiment without $L_\text{v}$ and compare results with our method.
It can be seen that there are many areas that should not exist above the reconstructed surface (see the 3rd column of Fig. \ref{ablation}), indicating that a relatively large spiking threshold is necessary, which is consistent with our analysis. 

\begin{table}[!t]
\Huge
    \centering
    \resizebox{1.03\hsize}{!}{
    \begingroup
    \fontsize{48}{53}\selectfont
    \begin{tabular}{l|c|c|c|c|c|c|c|c|c|c|c|c}
    \bottomrule
     & 37 & 40 & 55 & 63 & 65 & 83 & 97 & 105 & 110 & 114 & 118  & Avg. \\
        \midrule
        PET  & \textbf{0.84} & \textbf{0.72} & 0.40 & \textbf{0.95} & 0.84 & 1.42 & \textbf{1.14} & 0.72 & \textbf{1.05} & 0.44 & 0.59 &  \textbf{0.83} \\
        HFS  & 1.37 & 0.78 & 0.47 & 1.11 & \textbf{0.68} & 1.20 & 1.17 & 0.85 & 1.27 & 0.38 & 0.54  & 0.90 \\
        NeuS  & 1.37 & 0.93 & 0.43 & 1.10 & 0.70 & 1.48 & 1.16 & 0.83 & 1.69 & \textbf{0.35} & \textbf{0.49}  & 0.96 \\
        NeRF  & 1.63 & 1.75 & 0.60 & 2.03 & 1.07 & 1.70 & 1.95 & 1.13 & 2.33 & 0.87 & 1.05  & 1.46 \\
        Mip  & 1.83 & 1.65 & 1.61 & 2.90 & 1.79 & 1.93 & 2.19 & 1.32 & 2.52 & 1.38 & 1.49  & 1.87 \\
        Ours  & 1.20 & 1.02 & \textbf{0.38} & 1.15 & 0.72 & \textbf{1.10} & 1.19 & \textbf{0.65} & 1.60 & 0.49 & 0.55  & 0.90 \\
        \bottomrule
    \end{tabular}
    \endgroup
    }
  \caption{Quantitative Comparison on DTU \cite{80jensen2014large}. We show a quantitative comparison for the reconstructed geometry on 11 scans from the DTU dataset. 
  }
  \label{table1}
\end{table}

\paragraph{Effectiveness of B-FIF.}
To assess the effectiveness of B-FIF, we design an experiment that replaced B-FIF with FIF which does not have a maximum activation constraint, and compare the result with our method.
It can be seen that the reconstructed surface has significant deviation (see the 2nd column of Fig. \ref{ablation}), indicating that a bound is necessary, which is consistent with our analysis.
\paragraph{Effectiveness of smoothness.}
To assess the effectiveness of the initial smoothing, we design an experiment without $L_\text{g}$ and compare the result with our method.
It can be seen that although the Chamfer distances are similar, the reconstructed surface contains wrong high-frequency information (see the 4th column of Fig. \ref{ablation}), indicating that training from the wrong high-frequency surface is difficult. 
\section{Conclusion}
We present a novel neural surface reconstruction method for reconstructing objects.
NeRF and its variants use volume rendering to produce a neural scene representation. However, extracting high-quality surfaces from this learned implicit representation is difficult because there is not a definite interface density. In our work, we propose to use a hybrid ANN-SNN framework to reconstruct density fields. We observe that the conventional volume rendering method causes inherent geometric errors for surface reconstruction, and propose a new spiking neuron to get more accurate surface reconstruction with an adaptive spiking threshold.
\paragraph{Limitation.}
Since NeRF does not specifically consider scenes with high light and low brightness, it will struggle to accurately represent the geometry on these scenes \cite{11refnerf}. Our method also did not specifically consider these issues, so it is highly likely to struggle to accurately represent the geometry on these scenes. 

\section{Acknowledgements}
This work is supported by the National Natural Science Foundation of China under Grand No. 62376247. This work is supported by the State Key Lab of Brain-Machine Intelligence.

\bibliography{Formatting-Instructions-LaTeX-2024}

\begin{thebibliography}{60}
\providecommand{\natexlab}[1]{#1}

\bibitem[{Aimone et~al.(2022)Aimone, Date, Fonseca-Guerra, Hamilton, Henke, Kay, Kenyon, Kulkarni, Mniszewski, Parsa et~al.}]{aimone2022review}
Aimone, J.; Date, P.; Fonseca-Guerra, G.; Hamilton, K.; Henke, K.; Kay, B.; Kenyon, G.; Kulkarni, S.; Mniszewski, S.; Parsa, M.; et~al. 2022.
\newblock A Review of Non-Cognitive Applications for Neuromorphic Computing.
\newblock \emph{Neuromorphic Computing and Engineering}.

\bibitem[{Atzmon et~al.(2019)Atzmon, Haim, Yariv, Israelov, Maron, and Lipman}]{58atzmon2019controlling}
Atzmon, M.; Haim, N.; Yariv, L.; Israelov, O.; Maron, H.; and Lipman, Y. 2019.
\newblock Controlling Neural Level Sets.
\newblock \emph{NIPS}.

\bibitem[{Azinovi{\'c} et~al.(2022)Azinovi{\'c}, Martin-Brualla, Goldman, Nie{\ss}ner, and Thies}]{6RGB-Dsurfacereconstruction}
Azinovi{\'c}, D.; Martin-Brualla, R.; Goldman, D.~B.; Nie{\ss}ner, M.; and Thies, J. 2022.
\newblock Neural RGB-D Surface Reconstruction.
\newblock In \emph{CVPR}.

\bibitem[{Bagheri, Simeone, and Rajendran(2018)}]{25adversarial}
Bagheri, A.; Simeone, O.; and Rajendran, B. 2018.
\newblock Adversarial Training for Probabilistic Spiking Neural Networks.
\newblock In \emph{2018 IEEE 19th International Workshop on Signal Processing Advances in Wireless Communications (SPAWC)}.

\bibitem[{Barron et~al.(2021)Barron, Mildenhall, Tancik, Hedman, Martin-Brualla, and Srinivasan}]{8mipnerf}
Barron, J.~T.; Mildenhall, B.; Tancik, M.; Hedman, P.; Martin-Brualla, R.; and Srinivasan, P.~P. 2021.
\newblock Mip-NeRF: A Multiscale Representation for Anti-Aliasing Neural Radiance Fields.
\newblock In \emph{ICCV}.

\bibitem[{Boss et~al.(2021)Boss, Braun, Jampani, Barron, Liu, and Lensch}]{32boss2021nerd}
Boss, M.; Braun, R.; Jampani, V.; Barron, J.~T.; Liu, C.; and Lensch, H. 2021.
\newblock NeRD: Neural Reflectance Decomposition From Image Collections.
\newblock In \emph{ICCV}.

\bibitem[{Darmon et~al.(2022)Darmon, Bascle, Devaux, Monasse, and Aubry}]{darmon2022improving}
Darmon, F.; Bascle, B.; Devaux, J.-C.; Monasse, P.; and Aubry, M. 2022.
\newblock Improving Neural Implicit Surfaces Geometry with Patch Warping.
\newblock In \emph{CVPR}.

\bibitem[{Dayan and Abbott(2005)}]{dayan2005theoretical}
Dayan, P.; and Abbott, L.~F. 2005.
\newblock \emph{Theoretical Neuroscience: Computational and Mathematical Modeling of Neural Systems}.
\newblock MIT Press.

\bibitem[{Deng et~al.(2022)Deng, Liu, Zhu, and Ramanan}]{14Depth-supervisedNeRF}
Deng, K.; Liu, A.; Zhu, J.-Y.; and Ramanan, D. 2022.
\newblock Depth-Supervised NeRF: Fewer Views and Faster Training for Free.
\newblock In \emph{CVPR}.

\bibitem[{Fang et~al.(2021)Fang, Yu, Chen, Huang, Masquelier, and Tian}]{fang2021deep}
Fang, W.; Yu, Z.; Chen, Y.; Huang, T.; Masquelier, T.; and Tian, Y. 2021.
\newblock Deep Residual Learning in Spiking Neural Networks.
\newblock \emph{NIPS}.

\bibitem[{Fu et~al.(2022)Fu, Xu, Ong, and Tao}]{fu2022geo}
Fu, Q.; Xu, Q.; Ong, Y.~S.; and Tao, W. 2022.
\newblock Geo-Neus: Geometry-Consistent Neural Implicit Surfaces Learning for Multi-view Reconstruction.
\newblock \emph{NIPS}.

\bibitem[{Gehrig et~al.(2020)Gehrig, Shrestha, Mouritzen, and Scaramuzza}]{gehrig2020event}
Gehrig, M.; Shrestha, S.~B.; Mouritzen, D.; and Scaramuzza, D. 2020.
\newblock Event-Based Angular Velocity Regression with Spiking Networks.
\newblock In \emph{ICRA}.

\bibitem[{Guo et~al.(2022)Guo, Kang, Bao, He, and Zhang}]{29guo2022nerfren}
Guo, Y.-C.; Kang, D.; Bao, L.; He, Y.; and Zhang, S.-H. 2022.
\newblock NeRFReN: Neural Radiance Fields with Reflections.
\newblock In \emph{CVPR}.

\bibitem[{Han et~al.(2023)Han, Wang, Shen, and Tang}]{han2023symmetric}
Han, J.; Wang, Z.; Shen, J.; and Tang, H. 2023.
\newblock Symmetric-Threshold ReLU for Fast and Nearly Lossless ANN-SNN Conversion.
\newblock \emph{Machine Intelligence Research}.

\bibitem[{Ichnowski et~al.(2021)Ichnowski, Avigal, Kerr, and Goldberg}]{20dexnerf}
Ichnowski, J.; Avigal, Y.; Kerr, J.; and Goldberg, K. 2021.
\newblock Dex-NeRF: Using a Neural Radiance Field to Grasp Transparent Objects.
\newblock In \emph{5th Annual Conference on Robot Learning}.

\bibitem[{Jensen et~al.(2014)Jensen, Dahl, Vogiatzis, Tola, and Aan{\ae}s}]{80jensen2014large}
Jensen, R.; Dahl, A.; Vogiatzis, G.; Tola, E.; and Aan{\ae}s, H. 2014.
\newblock Large Scale Multi-View Stereopsis Evaluation.
\newblock In \emph{CVPR}.

\bibitem[{Kugele et~al.(2021)Kugele, Pfeil, Pfeiffer, and Chicca}]{kugele2021hybrid}
Kugele, A.; Pfeil, T.; Pfeiffer, M.; and Chicca, E. 2021.
\newblock Hybrid SNN-ANN: Energy-Efficient Classification and Object Detection for Event-Based Vision.
\newblock In \emph{DAGM German Conference on Pattern Recognition}.

\bibitem[{Lee et~al.(2020)Lee, Kosta, Zhu, Chaney, Daniilidis, and Roy}]{lee2020spike}
Lee, C.; Kosta, A.~K.; Zhu, A.~Z.; Chaney, K.; Daniilidis, K.; and Roy, K. 2020.
\newblock Spike-FlowNet: Event-Based Optical Flow Estimation with Energy-Efficient Hybrid Neural Networks.
\newblock In \emph{ECCV}.

\bibitem[{Lele et~al.(2022)Lele, Fang, Anwar, and Raychowdhury}]{lele2022bio}
Lele, A.~S.; Fang, Y.; Anwar, A.; and Raychowdhury, A. 2022.
\newblock Bio-Mimetic High-Speed Target Localization with Fused Frame and Event Vision for Edge Application.
\newblock \emph{Frontiers in Neuroscience}.

\bibitem[{Levy et~al.(2023)Levy, Peleg, Pearl, Rosenbaum, Akkaynak, Korman, and Treibitz}]{levy2023seathru}
Levy, D.; Peleg, A.; Pearl, N.; Rosenbaum, D.; Akkaynak, D.; Korman, S.; and Treibitz, T. 2023.
\newblock SeaThru-NeRF: Neural Radiance Fields in Scattering Media.
\newblock In \emph{CVPR}.

\bibitem[{Li et~al.(2021{\natexlab{a}})Li, Zhang, Mao, Xiao, Chang, and Zhou}]{18li2021fast}
Li, S.; Zhang, Z.; Mao, R.; Xiao, J.; Chang, L.; and Zhou, J. 2021{\natexlab{a}}.
\newblock A Fast and Energy-Efficient SNN Processor With Adaptive Clock/Event-Driven Computation Scheme and Online Learning.
\newblock \emph{IEEE Transactions on Circuits and Systems I: Regular Papers}.

\bibitem[{Li et~al.(2022)Li, Chen, Guo, Zhang, and Wang}]{21li2022brain}
Li, W.; Chen, H.; Guo, J.; Zhang, Z.; and Wang, Y. 2022.
\newblock Brain-Inspired Multilayer Perceptron with Spiking Neurons.
\newblock In \emph{CVPR}.

\bibitem[{Li et~al.(2021{\natexlab{b}})Li, Deng, Dong, Gong, and Gu}]{li2021free}
Li, Y.; Deng, S.; Dong, X.; Gong, R.; and Gu, S. 2021{\natexlab{b}}.
\newblock A Free Lunch from ANN: Towards Efficient, Accurate Spiking Neural Networks Calibration.
\newblock In \emph{ICML}.

\bibitem[{Li et~al.(2021{\natexlab{c}})Li, Guo, Zhang, Deng, Hai, and Gu}]{28li2021differentiable}
Li, Y.; Guo, Y.; Zhang, S.; Deng, S.; Hai, Y.; and Gu, S. 2021{\natexlab{c}}.
\newblock Differentiable Spike: Rethinking Gradient-Descent for Training Spiking Neural Networks.
\newblock \emph{NIPS}.

\bibitem[{Liu and Zhao(2022)}]{liu2022enhancing}
Liu, F.; and Zhao, R. 2022.
\newblock Enhancing Spiking Neural Networks with Hybrid Top-Down Attention.
\newblock \emph{Frontiers in Neuroscience}.

\bibitem[{Liu et~al.(2020{\natexlab{a}})Liu, Gu, Zaw~Lin, Chua, and Theobalt}]{68liu2020neural}
Liu, L.; Gu, J.; Zaw~Lin, K.; Chua, T.-S.; and Theobalt, C. 2020{\natexlab{a}}.
\newblock Neural Sparse Voxel Fields.
\newblock \emph{NIPS}.

\bibitem[{Liu et~al.(2020{\natexlab{b}})Liu, Zhang, Peng, Shi, Pollefeys, and Cui}]{69liu2020dist}
Liu, S.; Zhang, Y.; Peng, S.; Shi, B.; Pollefeys, M.; and Cui, Z. 2020{\natexlab{b}}.
\newblock DIST: Rendering Deep Implicit Signed Distance Function With Differentiable Sphere Tracing.
\newblock In \emph{CVPR}.

\bibitem[{Lu et~al.(2021)Lu, Jin, Pang, Zhang, and Karniadakis}]{lu2021learning}
Lu, L.; Jin, P.; Pang, G.; Zhang, Z.; and Karniadakis, G.~E. 2021.
\newblock Learning Nonlinear Operators via DeepONet Based on the Universal Approximation Theorem of Operators.
\newblock \emph{Nature Machine Intelligence}.

\bibitem[{Marchisio et~al.(2020)Marchisio, Nanfa, Khalid, Hanif, Martina, and Shafique}]{17isspikingsafe}
Marchisio, A.; Nanfa, G.; Khalid, F.; Hanif, M.~A.; Martina, M.; and Shafique, M. 2020.
\newblock Is Spiking Secure? A Comparative Study on the Security Vulnerabilities of Spiking and Deep Neural Networks.
\newblock In \emph{IJCNN}.

\bibitem[{Mildenhall et~al.(2021)Mildenhall, Srinivasan, Tancik, Barron, Ramamoorthi, and Ng}]{1nerf}
Mildenhall, B.; Srinivasan, P.~P.; Tancik, M.; Barron, J.~T.; Ramamoorthi, R.; and Ng, R. 2021.
\newblock NeRF: Representing Scenes as Neural Radiance Fields for View Synthesis.
\newblock \emph{Communications of the ACM}.

\bibitem[{Neftci, Mostafa, and Zenke(2019)}]{neftci2019surrogate}
Neftci, E.~O.; Mostafa, H.; and Zenke, F. 2019.
\newblock Surrogate Gradient Learning in Spiking Neural Networks: Bringing the Power of Gradient-Based Optimization to Spiking Neural Networks.
\newblock \emph{IEEE Signal Processing Magazine}.

\bibitem[{Niemeyer et~al.(2020)Niemeyer, Mescheder, Oechsle, and Geiger}]{71niemeyer2020differentiable}
Niemeyer, M.; Mescheder, L.; Oechsle, M.; and Geiger, A. 2020.
\newblock Differentiable Volumetric Rendering: Learning Implicit 3D Representations Without 3D Supervision.
\newblock In \emph{CVPR}.

\bibitem[{Oechsle, Peng, and Geiger(2021)}]{5unisurf}
Oechsle, M.; Peng, S.; and Geiger, A. 2021.
\newblock UNISURF: Unifying Neural Implicit Surfaces and Radiance Fields for Multi-View Reconstruction.
\newblock In \emph{ICCV}.

\bibitem[{Ororbia(2023)}]{ororbia2023spiking}
Ororbia, A. 2023.
\newblock Spiking Neural Predictive Coding for Continually Learning from Data Streams.
\newblock \emph{Neurocomputing}.

\bibitem[{Peng et~al.(2020)Peng, Niemeyer, Mescheder, Pollefeys, and Geiger}]{65peng2020convolutional}
Peng, S.; Niemeyer, M.; Mescheder, L.; Pollefeys, M.; and Geiger, A. 2020.
\newblock Convolutional Occupancy Networks.
\newblock In \emph{ECCV}.

\bibitem[{Peng et~al.(2021)Peng, Zhang, Xu, Wang, Shuai, Bao, and Zhou}]{30peng2021neuralbody}
Peng, S.; Zhang, Y.; Xu, Y.; Wang, Q.; Shuai, Q.; Bao, H.; and Zhou, X. 2021.
\newblock Neural Body: Implicit Neural Representations With Structured Latent Codes for Novel View Synthesis of Dynamic Humans.
\newblock In \emph{CVPR}.

\bibitem[{Pfeiffer and Pfeil(2018)}]{16Deeplearningwithspikingneurons}
Pfeiffer, M.; and Pfeil, T. 2018.
\newblock Deep Learning with Spiking Neurons: Opportunities and Challenges.
\newblock \emph{Frontiers in Neuroscience}.

\bibitem[{Pumarola et~al.(2021)Pumarola, Corona, Pons-Moll, and Moreno-Noguer}]{31dnerf}
Pumarola, A.; Corona, E.; Pons-Moll, G.; and Moreno-Noguer, F. 2021.
\newblock D-NeRF: Neural Radiance Fields for Dynamic Scenes.
\newblock In \emph{CVPR}.

\bibitem[{Ran{\c{c}}on et~al.(2022)Ran{\c{c}}on, Cuadrado-Anibarro, Cottereau, and Masquelier}]{ranccon2022stereospike}
Ran{\c{c}}on, U.; Cuadrado-Anibarro, J.; Cottereau, B.~R.; and Masquelier, T. 2022.
\newblock StereoSpike: Depth Learning With a Spiking Neural Network.
\newblock \emph{IEEE Access}.

\bibitem[{Ren et~al.(2023)Ren, Ma, Chen, Peng, Liu, Zhang, and Guo}]{ren2023spiking}
Ren, D.; Ma, Z.; Chen, Y.; Peng, W.; Liu, X.; Zhang, Y.; and Guo, Y. 2023.
\newblock Spiking PointNet: Spiking Neural Networks for Point Clouds.
\newblock In \emph{NIPS}.

\bibitem[{Saito et~al.(2019)Saito, Huang, Natsume, Morishima, Kanazawa, and Li}]{66saito2019pifu}
Saito, S.; Huang, Z.; Natsume, R.; Morishima, S.; Kanazawa, A.; and Li, H. 2019.
\newblock PIFu: Pixel-Aligned Implicit Function for High-Resolution Clothed Human Digitization.
\newblock In \emph{ICCV}.

\bibitem[{Schwarz et~al.(2020)Schwarz, Liao, Niemeyer, and Geiger}]{75schwarz2020graf}
Schwarz, K.; Liao, Y.; Niemeyer, M.; and Geiger, A. 2020.
\newblock GRAF: Generative Radiance Fields for 3D-Aware Image Synthesis.
\newblock \emph{NIPS}.

\bibitem[{Sharmin et~al.(2020)Sharmin, Rathi, Panda, and Roy}]{27inherentadversarial}
Sharmin, S.; Rathi, N.; Panda, P.; and Roy, K. 2020.
\newblock Inherent Adversarial Robustness of Deep Spiking Neural Networks: Effects of Discrete Input Encoding and Non-linear Activations.
\newblock In \emph{ECCV}.

\bibitem[{Shrestha and Orchard(2018)}]{shrestha2018slayer}
Shrestha, S.~B.; and Orchard, G. 2018.
\newblock SLAYER: Spike Layer Error Reassignment in Time.
\newblock \emph{NIPS}.

\bibitem[{Sitzmann, Zollh{\"o}fer, and Wetzstein(2019)}]{76sitzmann2019scene}
Sitzmann, V.; Zollh{\"o}fer, M.; and Wetzstein, G. 2019.
\newblock Scene Representation Networks: Continuous 3D-Structure-Aware Neural Scene Representations.
\newblock \emph{NIPS}.

\bibitem[{Verbin et~al.(2022)Verbin, Hedman, Mildenhall, Zickler, Barron, and Srinivasan}]{11refnerf}
Verbin, D.; Hedman, P.; Mildenhall, B.; Zickler, T.; Barron, J.~T.; and Srinivasan, P.~P. 2022.
\newblock Ref-NeRF: Structured View-Dependent Appearance for Neural Radiance Fields.
\newblock In \emph{CVPR}.

\bibitem[{Wang et~al.(2021)Wang, Liu, Liu, Theobalt, Komura, and Wang}]{7neus}
Wang, P.; Liu, L.; Liu, Y.; Theobalt, C.; Komura, T.; and Wang, W. 2021.
\newblock NeuS: Learning Neural Implicit Surfaces by Volume Rendering for Multi-view Reconstruction.
\newblock In \emph{NIPS}.

\bibitem[{Wang, Skorokhodov, and Wonka(2022)}]{wang2022hf}
Wang, Y.; Skorokhodov, I.; and Wonka, P. 2022.
\newblock HF-NeuS: Improved Surface Reconstruction Using High-Frequency Details.
\newblock \emph{NIPS}.

\bibitem[{Wang, Skorokhodov, and Wonka(2023)}]{wang2023pet}
Wang, Y.; Skorokhodov, I.; and Wonka, P. 2023.
\newblock PET-NeuS: Positional Encoding Tri-Planes for Neural Surfaces.
\newblock In \emph{CVPR}.

\bibitem[{Xu et~al.(2019)Xu, Wang, Ceylan, Mech, and Neumann}]{67xu2019disn}
Xu, Q.; Wang, W.; Ceylan, D.; Mech, R.; and Neumann, U. 2019.
\newblock DISN: Deep Implicit Surface Network for High-quality Single-view 3D Reconstruction.
\newblock \emph{NIPS}.

\bibitem[{Yariv et~al.(2021)Yariv, Gu, Kasten, and Lipman}]{yariv2021volume}
Yariv, L.; Gu, J.; Kasten, Y.; and Lipman, Y. 2021.
\newblock Volume Rendering of Neural Implicit Surfaces.
\newblock \emph{NIPS}.

\bibitem[{Yariv et~al.(2020)Yariv, Kasten, Moran, Galun, Atzmon, Ronen, and Lipman}]{79yariv2020multiview}
Yariv, L.; Kasten, Y.; Moran, D.; Galun, M.; Atzmon, M.; Ronen, B.; and Lipman, Y. 2020.
\newblock Multiview Neural Surface Reconstruction by Disentangling Geometry and Appearance.
\newblock \emph{NIPS}.

\bibitem[{Zhang, Huang, and He(2023)}]{zhang2023defects}
Zhang, C.; Huang, C.; and He, J. 2023.
\newblock Defects Recognition of Train Wheelset Tread Based on Improved Spiking Neural Network.
\newblock \emph{Chinese Journal of Electronics}.

\bibitem[{Zhang et~al.(2022)Zhang, Dong, Zhang, Ding, Heide, Yin, and Yang}]{22SpikingTransformers}
Zhang, J.; Dong, B.; Zhang, H.; Ding, J.; Heide, F.; Yin, B.; and Yang, X. 2022.
\newblock Spiking Transformers for Event-Based Single Object Tracking.
\newblock In \emph{CVPR}.

\bibitem[{Zhang et~al.(2023)Zhang, Huang, Ma, and Zhou}]{zhang2023predicting}
Zhang, J.; Huang, L.; Ma, Z.; and Zhou, H. 2023.
\newblock Predicting the Temporal-Dynamic Trajectories of Cortical Neuronal Responses in Non-Human Primates Based on Deep Spiking Neural Network.
\newblock \emph{Cognitive Neurodynamics}.

\bibitem[{Zhang et~al.(2020)Zhang, Riegler, Snavely, and Koltun}]{2nerf++}
Zhang, K.; Riegler, G.; Snavely, N.; and Koltun, V. 2020.
\newblock NeRF++: Analyzing and Improving Neural Radiance Fields.
\newblock \emph{arXiv preprint arXiv:2010.07492}.

\bibitem[{Zhang et~al.(2021)Zhang, Srinivasan, Deng, Debevec, Freeman, and Barron}]{12nerfactor}
Zhang, X.; Srinivasan, P.~P.; Deng, B.; Debevec, P.; Freeman, W.~T.; and Barron, J.~T. 2021.
\newblock NeRFactor: Neural Factorization of Shape and Reflectance Under an Unknown Illumination.
\newblock \emph{TOG}.

\bibitem[{Zhao et~al.(2022)Zhao, Yang, Zheng, Wu, Liu, Wu, Li, Chen, Song, Zhu et~al.}]{zhao2022framework}
Zhao, R.; Yang, Z.; Zheng, H.; Wu, Y.; Liu, F.; Wu, Z.; Li, L.; Chen, F.; Song, S.; Zhu, J.; et~al. 2022.
\newblock A Framework for the General Design and Computation of Hybrid Neural Networks.
\newblock \emph{Nature Communications}.

\bibitem[{Zhu et~al.(2022)Zhu, Wang, Chang, Li, Huang, and Tian}]{23zhu2022event}
Zhu, L.; Wang, X.; Chang, Y.; Li, J.; Huang, T.; and Tian, Y. 2022.
\newblock Event-Based Video Reconstruction via Potential-Assisted Spiking Neural Network.
\newblock In \emph{CVPR}.

\bibitem[{Zou, Huang, and Wu(2022)}]{zou2022towards}
Zou, X.-L.; Huang, T.-J.; and Wu, S. 2022.
\newblock Towards a New Paradigm for Brain-Inspired Computer Vision.
\newblock \emph{Machine Intelligence Research}.

\end{thebibliography}



\newpage
\twocolumn[
\centering
\Large
\textbf{Supplementary Material for ``Spiking NeRF: Representing \\the Real-World Geometry by a Discontinuous Representation"}\\
\vspace{1.0em}
]

This supplementary material consists of two parts, the additional analysis and the additional results.

\begin{figure}[!t]

\centering
\includegraphics[width=1\columnwidth]{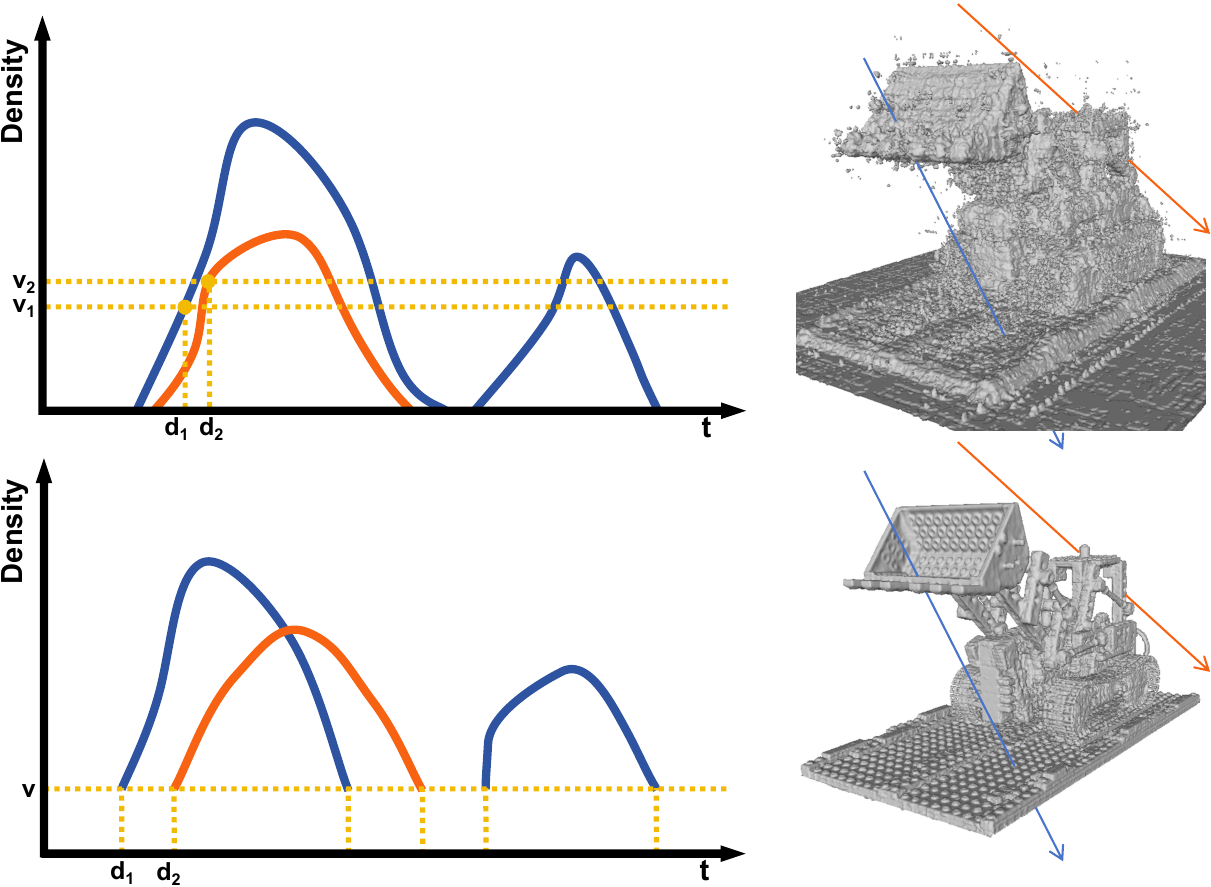}
\caption{
The blue line and orange line in the right represent two different light. The figures on the left represent the distribution of the density field through which light passes. $t$ is the sampling distance from the sampling point to the camera origin.
$d_1$ and $d_2$ represent $t$ corresponding to the position of surface points along the direction of light rays (i.e., the depth corresponding to the direction of light). $v_1$ and $v_2$ represent the density corresponding to the position of surface points along the direction of light rays.
For continuous fields, using the same threshold to extract geometric information cannot guarantee that the depth corresponding to the light is all accurate.
}
\label{light}
\end{figure}
\section{Additional Analysis}
\subsection{Optimal Threshold Perturbation}
For a well-trained NeRF, $\int_0^T {\sigma(t)\cdot\exp\left({-\int_0^t {\sigma(t)} \,{\rm d}t}\right)\cdot t} \,{\rm d}t$ can represent the depth of the corresponding light direction \cite{12nerfactor,29guo2022nerfren}. However, it depends on the direction. The final geometric information is extracted from the network using a threshold, independent of the direction. 
As shown in Fig. \ref{light}, when light from different directions passes through a density field, the corresponding density distribution is different, resulting in different density values corresponding to depth. So using the same threshold to extract geometric information cannot guarantee that the depth corresponding to the light is all accurate.






\subsection{The Proof of Proposition 1}
Define $m$ is the index of the first non zero $\sigma$ (i.e., the extracted depth $d_\text{v} = t_m$), and $m'$ is the index of the largest $\sigma$ within m+1 and N (i.e., $V_{max}=\sigma_{m'}$). $\Delta t_i$ is the sampling interval between sampling point $i-1$ and point $i$.  The $w_i$ refers to the weight at each sampling point as defined in \cite{1nerf}.
We define $\beta_i = e^{-\sigma_i \Delta t_i}$. Then we have:
\begin{equation}
        \sum_i^N w_i = 1 - \prod_i^N \beta_i.
\end{equation}
Let $c = \prod_{i=1}^{N} \beta_i$. Since $\sigma_1$, ..., $\sigma_{m-1}$ are all 0, we have:
\begin{equation}
    \begin{aligned}
        0<\beta_{m}e^{-\sigma_{m'}(t_N-t_{m})}<c< \beta_{m}\beta_{m'}
    \end{aligned}
\end{equation}

According to previous analysis, for well trained NeRF, it can be considered that the depth $d$ is $\int_0^T {\sigma(t)\cdot\exp\left({-\int_o^t {\sigma(t)} \,{\rm d}t}\right)\cdot t} \,{\rm d}t$. Its Riemann sum form is $\sum_{i=1}^{N}{w_i t_i}$.
\begin{equation}
    \begin{aligned}
        d - t_m &= \sum_{i=1}^{N}{w_i t_i} - t_m\\
        &= \sum_{i=m}^{N}{w_i t_i} - t_m\left(\sum_{i=1}^{N}{w_i}+c\right)\\
        &\ge \left(1-w_m-c\right)\left(t_{m+1}-t_m\right) - t_m c\\
        &\ge \left(t_{m+1}-t_m\right) \beta_{m} - t_{m+1} \beta_{m}\beta_{m'}\\
        &\ge \left(\Delta t_{m+1} - Te^{-\sigma_{m'}\Delta t_{m'}}\right) e^{-\sigma_{m}\Delta t_{m}}.
    \end{aligned}
\end{equation}
For a well-trained NeRF, the spiking threshold $V_\text{th}$ equals to the non zero minimum value of its density field. Because the value before the last layer varies continuously with respect to the spatial point, the density value of the first point that a ray encounters with a non zero density equals to the non zero minimum value of its density field. Then $V_\text{th}=\sigma_m$. So we have:

\begin{equation}
    \begin{aligned}
        d - t_m &\ge \left(\Delta t_{m+1} - Te^{-\sigma_{m'}\Delta t_{m'}}\right) e^{-\sigma_{m}\Delta t_{m}}\\
        &= \left(\Delta t_{m+1} - Te^{-V_\text{max}\Delta t_{m'}}\right) e^{-V_\text{th}\Delta t_{m}}.\\
    \end{aligned}
\end{equation}
\begin{equation}
    \begin{aligned}
        d - t_m &= \sum_{i=1}^{N}{w_i t_i} - t_m\\
        &= \sum_{i=m}^{N}{w_i t_i} - t_m\left(\sum_{i=1}^{N}{w_i}+c\right)\\
        &\le \left(t_{N}-t_m\right)\left(1-w_m-c\right) - t_m c\\
        &\le \left(t_{N}-t_m\right) \beta_{m} - t_{N} \beta_{m}e^{-\sigma_{m'}\left(t_N-t_{m}\right)}\\
        &\le \left(t_{N}-t_{N} e^{-\sigma_{m'}t_N}\right)\beta_{m}\\
        &= T\left(1-e^{-V_\text{max}T}\right)e^{-V_\text{th}\Delta t_{m}}.\\
    \end{aligned}
\end{equation}
Since we do not explore the influence of sampling point positions, we have written $\Delta t_{m}$,$\Delta t_{m+1}$ and $\Delta t_{m'}$ in the equation as $\Delta t$ in the paper for the convenience of observation.

Then we have:
\begin{equation}
    \resizebox{.89\hsize}{!}{$\left(\Delta t - Te^{-V_\text{max}\Delta t}\right) e^{-V_\text{th}\Delta t}<d - d_\text{v}<T\left(1-e^{-V_\text{max}T}\right)e^{-V_\text{th}\Delta t}$}.
\end{equation}

\begin{figure}[t]
\centering
\includegraphics[width=1\columnwidth]{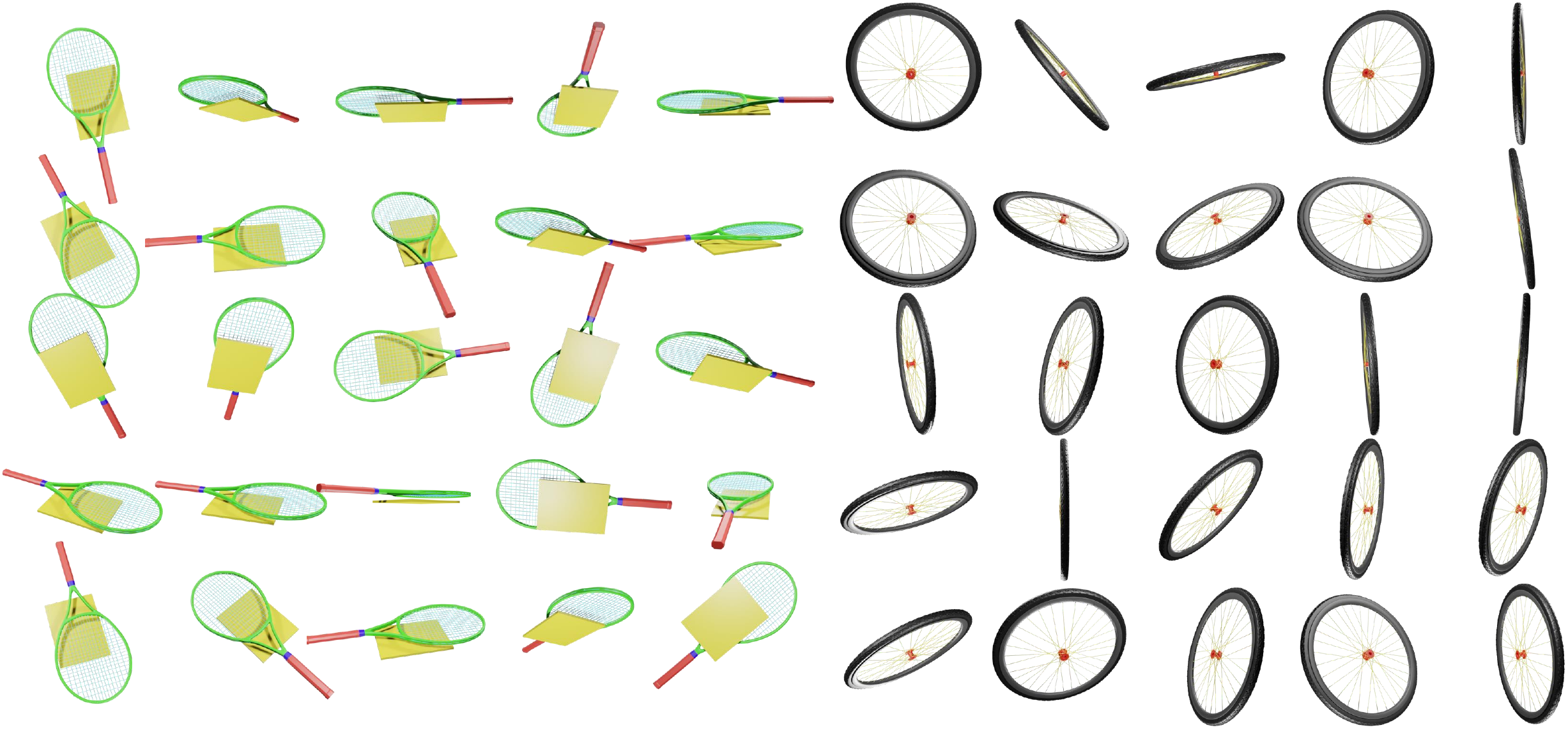}
\caption{Self created scenes. Left: A tennis racket with a board. Right: A wheel.
}
\label{data}
\end{figure}

For a general setting, T is greater than 100 times $\Delta t$. 
And $\Delta t$ is approximately $0.01$.
So $\left|\Delta t - Te^{-V_\text{max}\Delta t}\right| <= T\left(1-e^{-V_\text{max}T}\right)$ for $V_\text{max}>4$.
Therefore, for a fixed $V_\text{th}>4$, which is basically valid, a small $V_\text{max}$ can decrease the error.
\begin{figure}[!t]
\centering
\includegraphics[width=0.92\columnwidth]{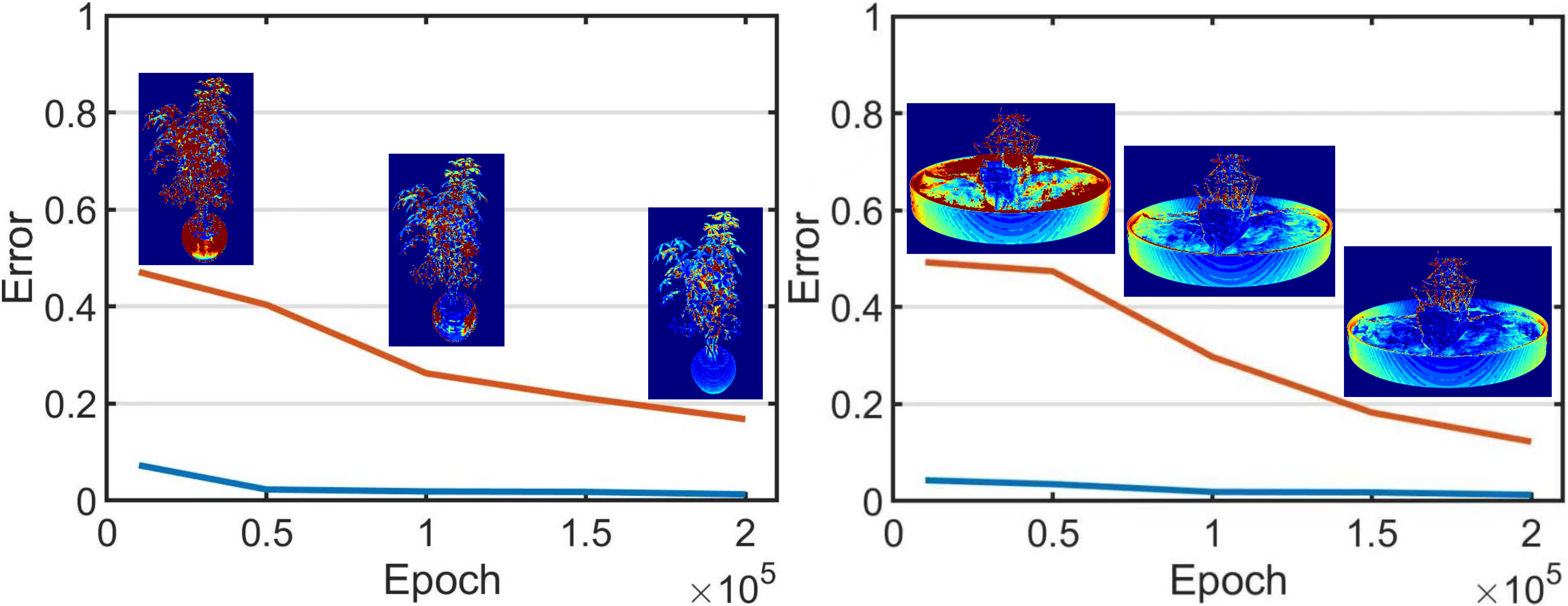}
\caption{The relationship between the upper bound and the average depth error during training. We show two more scenes from Blender dataset \cite{1nerf}. The red curve represents upper bound while the blue curve represents the average depth error during training. 
}
\label{bound_2}
\end{figure}



\subsection{Other Bounded Functions}
\paragraph{IF spiking neuron.}
The IF spiking neuron has a bound. However, according to the analysis in \textbf{Preliminary}, it is ultimately difficult to maintain accuracy using IF spiking neuron.
\paragraph{FIF with a hard bound.}
\begin{equation}
    \label{eq:55}
    o_{t+1} = 
    \begin{cases}
        0&u^\text{pre}_{t+1}<V_\text{th} \\
        1&\text{otherwise},
    \end{cases}
\end{equation}
\begin{equation}
    \label{eq:66}
    y_{t+1} = \min(o_{t+1} \cdot V_\text{th}, B).
\end{equation}
Here, $B$ is the upper bound of FIF spiking neuron. Although there is a bound, increasing the bound requires additional operations (e.g., additional loss term or a strategy to increase $B$), If $B$ is not updated, the network will only output $0$ when $V_\text{th}$ exceeds $B$. 

We use $k\tanh()$ in the paper because $k$ can be updated through the original loss. 
\subsection{Analysis of Surrogate Gradient}
The surrogate gradient in \cite{21li2022brain} is:
\begin{equation}
    \label{eq:162}
    {{\partial L}\over{\partial V_\text{th}}} = \sum{{{\partial L}\over{\partial y_t}} (1-o_t)}.
\end{equation}
Due to $t=1$ in our paper. We have:
\begin{equation}
    \label{eq:161}
    {{\partial L}\over{\partial V_\text{th}}} = {{{\partial L}\over{\partial y}} (1-o)}.
\end{equation}
Since the value of $1-o$ can only be $0$ or $1$, it can lead to unstable training.

\section{Additional Results}
\subsection{Introduction of Our Dataset}
We additionally generate our own dataset containing pathtraced images
of two objects. They are all rendered from viewpoints uniformly sampled on a full sphere. We render 100 views of each scene as input and 200 for testing, all at 800 × 800 pixels (see Fig. \ref{data}).


\subsection{Validation for Upper Bound of Proposition 1}
We show two more scenes for the upper bound of proposition 1 in Fig. \ref{bound_2}. It can be seen that the average depth error decreases with the upper bound and the average depth error keeps being less than upper bound during training.

\end{document}